%% file: main.tex
\documentclass[10pt, conference, letterpaper]{IEEEtran}

\usepackage{graphicx}
\usepackage{caption,subcaption}
\usepackage{amsmath}
\usepackage{amsfonts}
\usepackage[ruled]{algorithm2e}
\usepackage[noend]{algpseudocode}
\usepackage{url}
\usepackage{multirow}
\usepackage{cite}
\newcommand{\argmax}{\mathop{\rm arg~max}\limits}
\newcommand{\argmin}{\mathop{\rm arg~min}\limits}

\newtheorem{proposition}{Proposition}

\newtheorem{lemma}{Lemma}

\graphicspath{ {figures/} {diagrams/} {individuals/}}

\begin{document}

\title{MOVI: A Model-Free Approach \\ to Dynamic Fleet Management}

\author{
\IEEEauthorblockN{Takuma Oda and Carlee Joe-Wong}
\IEEEauthorblockA{Carnegie Mellon University \\
takumao@andrew.cmu.edu, cjoewong@andrew.cmu.edu
}
}
\date{Today}
\maketitle

\begin{abstract}
Modern vehicle fleets, e.g., for ridesharing platforms and taxi companies, can reduce passengers' waiting times by proactively dispatching vehicles to locations where pickup requests are anticipated in the future. Yet it is unclear how to best do this: optimal dispatching requires optimizing over several sources of uncertainty, including vehicles' travel times to their dispatched locations, as well as coordinating between vehicles so that they do not attempt to pick up the same passenger. While prior works have developed models for this uncertainty and used them to optimize dispatch policies, in this work we introduce a model-free approach. Specifically, we propose MOVI, a Deep Q-network (DQN)-based framework that directly learns the optimal vehicle dispatch policy. Since DQNs scale poorly with a large number of possible dispatches, we streamline our DQN training and suppose that each individual vehicle independently learns its own optimal policy, ensuring scalability at the cost of less coordination between vehicles.
We then formulate a centralized receding-horizon control (RHC) policy to compare with our DQN policies. To compare these policies, we design and build MOVI as a large-scale realistic simulator based on 15 million taxi trip records that simulates policy-agnostic responses to dispatch decisions. We show that the DQN dispatch policy reduces the number of unserviced requests by 76\% compared to without dispatch and 20\% compared to the RHC approach, emphasizing the benefits of a model-free approach and suggesting that there is limited value to coordinating vehicle actions. This finding may help to explain the success of ridesharing platforms, for which drivers make individual decisions.
\end{abstract}

\section{Introduction}\label{sec:introduction}
\input{introduction.tex}

\section{Problem Definition}\label{sec:definition}
\input{definition.tex}

\section{RHC Policy Baseline}\label{sec:RHC}
\input{RHC.tex}

\section{Distributed DQN Policy}\label{sec:DQN}
\input{DQN.tex}

\section{MOVI Fleet Simulator Design}\label{sec:experiments}
\input{experiments.tex}

\section{Results and Discussion}\label{sec:results}
\input{results.tex}

\section{Conclusion}\label{sec:conclusion}
\input{conclusions.tex}

\input{reference.tex}
\clearpage

\end{document}

%% file: introduction.tex
With the development of smart devices and large-scale data processing technology, most ride-hailing fleet networks (e.g., Uber, Lyft, and taxi services) can now track vehicles' GPS locations and passengers' pickup requests in real time. This data can then be utilized to predict passenger demand and vehicle mobility patterns in the future, reducing passengers' waiting times by proactively dispatching vehicles to predicted future pickup locations \cite{RHC}.

Proactive taxi\footnote{We use ``taxi'' and ``vehicle'' interchangeably in this work.} dispatch over a large city poses significant \emph{coordination} and \emph{uncertainty} challenges: it requires real-time decision making over uncertain future demand for thousands of drivers competing to service pickup requests. Moreover, individual drivers may have an incentive to deviate from coordinated solutions, e.g., if the globally optimal coordinated solution requires them to drive a long distance. Solving these challenges simultaneously is difficult: computing a coordinated dispatch solution for thousands of vehicles may take time, exacerbating the uncertainty challenge of optimizing over rapidly changing passenger demands. Even evaluating possible solutions is difficult due to the many sources of future uncertainty (e.g., passenger demand, vehicle trip times), which are hard to model. Yet realistic models are needed to assess the tradeoffs between multiple, possibly conflicting objectives like minimizing the passenger waiting time, the number of unserved requests, and vehicles' idle cruising time. For instance, vehicles may need to drive long distances to the locations with predicted pickup requests, increasing their idle cruising time to reduce the number of unserved requests. Thus, in this work we answer two major research questions:
\begin{itemize}
\item
\emph{Can a distributed dispatch approach that does not rely on system models outperform a coordinated approach?}
\item
\emph{What are the performance tradeoffs of these approaches in a realistic environment with uncertain future demand, vehicle trip times, and driving routes?}
\end{itemize}

\subsection{Related Work}

Traditional taxi networks dispatch taxis by having individual drivers look for passengers hailing vehicles on the street. Digitizing these systems allows drivers to view passenger demands through a mobile application and move to regions of higher demand, reducing passenger waiting times. However, such apps still rely on drivers' human intuition; they do not show \emph{future} demand, preventing drivers from proactively heading to locations where future pickups are likely. Our goal is to develop \emph{optimized dispatch algorithms} that do not rely on human intuition and account for likely future demands.

Most previous works on fleet management address prediction challenges with a model-based approach, which first models pickup request locations, vehicle travel times, etc. and then optimally dispatches vehicles given these models. Indeed, vehicle routing from a central depot is a classical operations research problem~\cite{laporte1992vehicle,golden1998impact,cordeau1997tabu}. Recent studies have taken advantage of real-time taxi information to fit system models and minimize passengers' waiting times and vehicle cruising times~\cite{wait, demand1, demand2, jauhri}. For instance, Miao et.al \cite{RHC,miao2017data} designed a Receding Horizon Control (RHC) framework, which incorporates a demand/supply model and real-time GPS location and occupancy information. Both studies show a reduction in the total idle distance through extensive trace-driven analysis. Others have proposed matching algorithms~\cite{zheng2017online} and re-balancing methods for autonomous vehicles~\cite{robotic}, considering both global service fairness and future costs.

Though the model-based approaches considered in these works can improve system performance, they are inherently limited by pre-specifying system models~\cite{kaelbling1996reinforcement}. Such specification may be particularly restrictive in a highly dynamic environment like fleet management, where components like trip times and the actual routes vehicles should take must be continually updated based on historical information. 
%

In this work, we introduce MOVI (Model-free Optimization of Vehicle dIspatching), the first \emph{model-free} approach to fleet management. MOVI uses a reinforcement learning technique called deep Q-network (DQN)~\cite{DL,DQN} that focuses on finding the optimal actions rather than accurately modeling the system. DQNs' known strengths for systems with a large number of input variables allow them to solve the uncertainty challenge presented by fleet management, but they exacerbate the coordination challenge: the complexity of the DQN solution grows exponentially with the number of dispatch possibilities, which in our scenario can be very large given the thousands of taxi vehicles in a city. Indeed, most model-free approaches would face this challenge, due to their lack of a model to guide the search through dispatch possibilities. Thus, we take a \emph{distributed approach} in which each vehicle solves its own DQN problem, without coordination. We introduce a new DQN training method to ensure fast training at each vehicle.
 
Prior studies have taken a similar vehicle-centric approach by providing route recommendations that aim to maximize individual drivers' profits~\cite{recommend1, recommend2} or modeling individual driver behavior~\cite{ziebart2008navigate}. We show that a distributed DQN decision framework outperforms a model-based centralized dispatch framework, indicating that model-free approaches can add significant value to fleet management and that there may be limited value to a coordinated vehicle dispatch approach.

\subsection{Our Contributions}
In this paper, we focus on modern fleet networks that can collect vehicles' GPS location and occupancy status in real time and receive pickup requests from passengers over the Internet at a cloud-based dispatch center. Our goal is to optimally direct a fleet of taxi vehicles to different locations in a city so as to minimize passengers' wait times and vehicles' idle driving costs. 
Our contributions are as follows:
\begin{itemize}
\item To the best of our knowledge, \textbf{MOVI is the first work to design a model-free approach} for a large-scale taxi dispatch problem. To ensure scalability, we use a distributed DQN with streamlined training algorithm.

\item To evaluate our model-free, distributed DQN approach, we formulate a baseline {\bf model-based, centralized RHC policy} based on a linear program, integrating predicted demands and trip times and fleet system dynamics.

\item We \textbf{design and build MOVI as a large-scale realistic fleet simulator} based on 15.6 million New York City taxi records and Open Street Map road data \cite{osm, data}. MOVI uses a modular architecture that ensures policy-agnostic dispatch responses from the simulated environment, allowing us to fairly compare our RHC and DQN policies. 

\item In spite of relying on individual decisions, we find that \textbf{our DQN approach reduces the average reject rate
by 76\%} compared to the results without dispatch and by 20\% compared to the model-based RHC approach in our simulator. Moreover, DQN leads to a higher minimum vehicle utilization rate, indicating that drivers have more incentive to follow its policies.

\end{itemize}
A DQN-based dispatch framework not only outperforms RHC, but also offers significant practical benefits, e.g., being more scalable to large numbers of drivers. We formally define the taxi dispatch problem in Section~\ref{sec:definition} before introducing our RHC and DQN policies in Sections~\ref{sec:RHC} and~\ref{sec:DQN} respectively. We then present our fleet simulator in Section~\ref{sec:experiments} and our results in Section~\ref{sec:results}. We finally conclude the paper in Section~\ref{sec:conclusion}.

%% file: definition.tex
\begin{figure}
\centering
\includegraphics[width=0.43\textwidth]{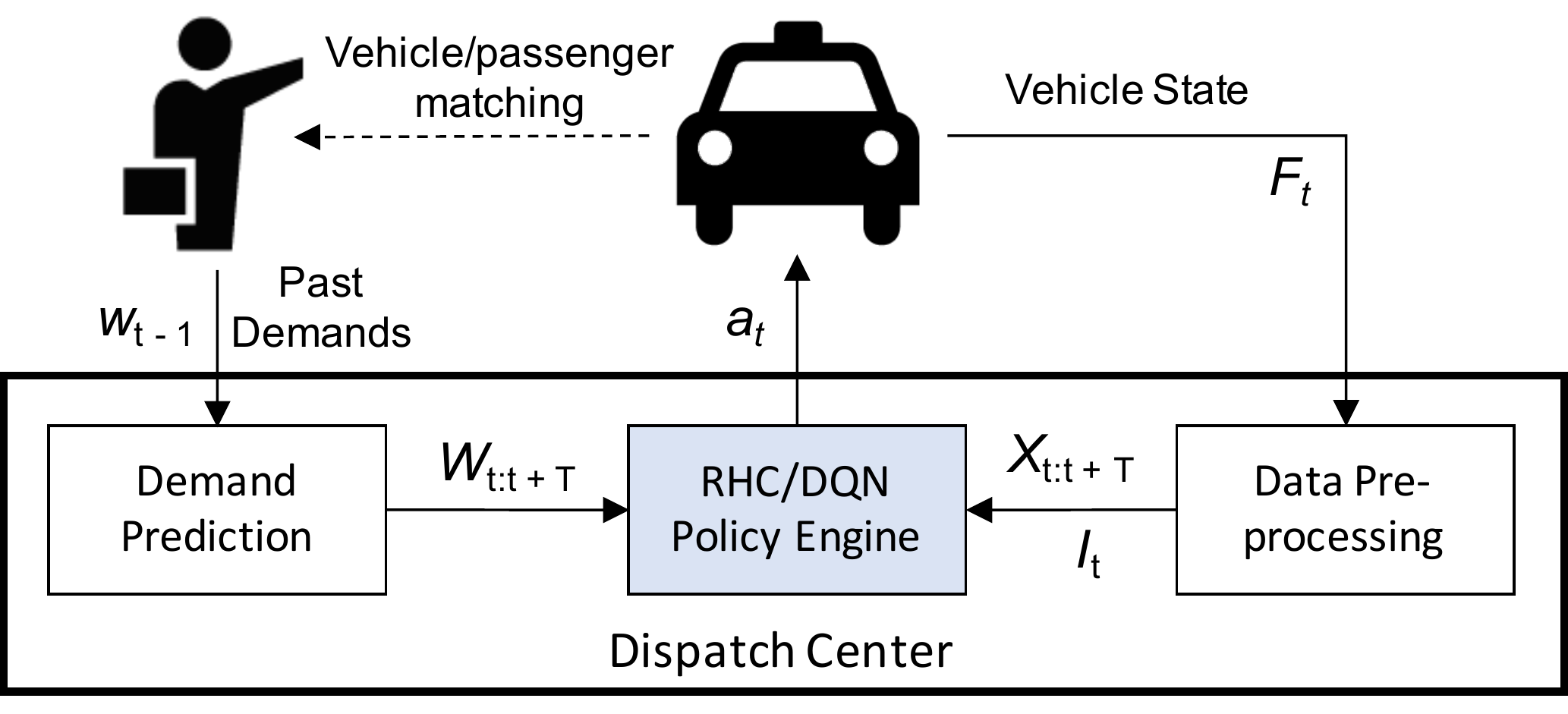}
\caption{Interaction of vehicles and passengers with the dispatch center. Our dispatch policies compute an action $a_t$ based on the environment state $s_t = \left((F_t, X_{t(t:t+T)}, W_{t:t+T}\right)$.}
\label{fig:agent}
\end{figure}

We assume the ride service consists of a dispatch center, a large number of geographically distributed vehicles, and passengers with a mobile ride request application. Figure~\ref{fig:agent} illustrates this framework. The dispatch center tracks each vehicle's real-time GPS location and availability status and all passenger pickup requests. It uses this information to proactively dispatch vehicles to locations where it predicts future pickups will be requested, and to match vehicles to incoming pickup requests. We focus our optimization on policies for proactive dispatching, as shown in Figure~\ref{fig:agent}, rather than vehicle matching. In this section, we formulate the proactive dispatch problem using the notation summarized in Table~\ref{table:params}.

We view the dispatch center as an agent that interacts with its external environment through a sequence of observations, actions and rewards. We divide the geographical service area into $M$ regions 
and consider $T$ timeslots of length $\Delta t$ indexed by $t = t_0 + 1,\ldots,t_0 + T$, where $t_0$ is the current timeslot. The number of pickup requests at the $i$-th region within time slot $t$ is then denoted by $w_{t,i}$, and the number of available vehicles in this region at the beginning of time slot $t$ is denoted by $x_{t,i}$. We also define $x_{tt', i}$ as the number of vehicles that are occupied at time t but will drop off passengers and become idle in the $i$-th region in time slot $t'$. To predict the future $x_{tt', i}$ given a set of dispatch actions, we use $F_t = (f_t^{(1)}, \dots, f_t^{(N)})$ to denote the current location, occupied/idle status and destination of each vehicle available at time $t$ for the dispatch center. By combining this data, we can predict $X_{t(t:t+T)} = (x_t, \ldots, x_{t+T})$, a matrix that gives the number of vehicles available in each region from time $t$ to time $t+T$, given the dispatch actions. Similarly, we define the future demand $W_{t:t+T} = (\bar{w}_{t}, \ldots, \bar{w}_{t+T})$. The state $s_t$ of the external environment at time $t$ is then $s_t = (F_t, X_{t(t:t+T)}, W_{t:t+T})$. 

At each time step $t$, the agent receives some representation of the environment's state $s_t$ and reward $r_t$. It then takes action $a_t$ to dispatch vehicles to the different regions so as to maximize the expected future reward:
\begin{equation}
\label{eqn:reward}
\sum_{t'=t}^{\infty} \gamma ^{t'-t} r_{t'}(a_t, s_t)
\end{equation}
where $\gamma < 1$ represents a time discount rate. The action $a_t$ routes idle vehicles (i.e., with $f_t^{(i)} = 1$), the set of which we denote by $\mathcal{I}_t$, to different regions. We formally define $a_t$ and $r_t$ for each policy in Sections~\ref{sec:RHC} and~\ref{sec:DQN}.  To define $r_t$, we wish to minimize three performance criteria: the number of service rejects, passenger waiting time and idle cruising time. A \emph{reject} means a ride request that could not be served within a given amount of time because of no available vehicles near a customer. The \emph{waiting time} is defined by the time between a passenger's placing a pickup request and the matched driver picking up the passenger; even if a request is not rejected, passengers would prefer to be picked up sooner rather than later. Finally, the \emph{idle cruising time} is the time in which a taxi is unoccupied and therefore not generating revenue, while still incurring costs like gasoline and wear on the vehicle. 
%


In the next two sections, we develop a baseline Receding Horizon Control (RHC) policy and a Deep Q-Network (DQN) policy to solve this dispatch problem.

\begin{table}[t]
\caption{Notation used in the RHC and DQN formulations.}
\vspace{-0.05in}
\label{table:params}
\centering
\begin{tabular}{|l|l|} \hline
{\bf Parameters} & {\bf Description} \\ \hline
$N$ & the number of vehicles \\ \hline
$M$ & the number of regions \\ \hline
$\gamma \in (0,1]$ & time discount rate \\ \hline
$\Delta t$ & step size \\ \hline
$T$ & maximum time steps \\ \hline

$s_t$ & state of the environment at the beginning of $t$ \\ \hline
$a_t$ & action taken at the beginning of $t$ (dispatch order) \\ \hline
$r_t$ & reward gained at the beginning of $t$ \\ \hline
$f_t^{(n)}$ & $n$-th vehicle's state at the beginning of $t$ \\ \hline

$x_t \in \mathbb{Z} ^{M}$ & number of idle vehicles in each region at time slot $t$ \\ \hline
\multirow{2}{*}{$x_{tt'} \in \mathbb{Z} ^{M}$} & number of occupied vehicles at time $t$ that become \\
& idle at time $t'$\\ \hline
$w_t \in \mathbb{Z} ^{M}$ & number of requests in each region at time slot $t$\\ \hline
\multirow{2}{*}{$\bar{w}_t \in \mathbb{Z} ^{M}$} & number of predicted requests in each region \\
& at time slot $t$\\ \hline
\multirow{2}{*}{$u_t\in \mathbb{Z} ^{M \times M}$} & number of vehicles to be dispatched between regions \\
& at time slot t\\ \hline
$\tau_{t} \in \mathbb{R} ^{N \times N}$ & expected travel time between the regions at time slot t\\ \hline
\multirow{2}{*}{$\mathbb{P}_t(d|o)$} & probability distribution of the destination region $d$ \\
& given the origin region $o$ at time slot $t$\\ \hline
$\lambda$ & cost of a reject \\ \hline
$\theta$ & network parameters in Q-network ($Q$)\\ \hline
$\theta ^-$ & network parameters in target-network ($\bar{Q})$ \\ \hline
$\eta(l)$ & demand supply distribution mismatch \\ \hline

\end{tabular}
\end{table}

%% file: RHC.tex
In the RHC formulation, we define our action variables $a_t$ in Section~\ref{sec:definition} to be $u_{t}\in\mathbb{Z}^{M\times M}$, where each $u_{t,ij}$ is the number of vehicles dispatched within time slot $t$ from the $i$-th to the $j$-th region. We wish to choose the $u_t$ so as to minimize a weighted sum of the number of rejects and the vehicles' idle cruising time, defining the reward as the negative of this sum:
\begin{equation}
r_t(u_t) = -\lambda \sum_{i=1}^M \text{min}(x_{t,i}-\bar{w}_{t,i}, 0) - \sum_{i, j =1}^M \tau_{t,ij} u_{t,ij}
\label{eq:reward_RHC}
\end{equation}
The first term in this objective, $\text{min}(x_{t,i}-\bar{w}_{t,i}, 0)$, represents the difference between taxi demand and supply ($x_{t,i} - \bar{w}_{t,i}$) at each region $i$ within time slot $t$. Demand that cannot be served by these resources is deemed rejected\footnote{This definition can be easily extended by summing over multiple time slots $t$ in $\min\left(x_{t,i} - \bar{w}_{t,i},0\right)$ to allow for greater waiting times before rejection.}. The second term in (\ref{eq:reward_RHC}) corresponds to the idle vehicle cruising cost, where $\tau_{t,ij}$ is the expected travel time from the $i$-th to the $j$-th region. Here $\lambda$ weights the rejection cost compared to the idle cruising time.

To find $x_{t,i} - \bar{w}_{t,i}$ in terms of the action variables $u_t$, we find the future number of available vehicles:
\begin{lemma}
The number of idle vehicles in each time slot is:
\begin{align}
x_{t+1, i} = & \text{max}(x_{t,i}-\bar{w}_{t+1,i}, 0) - \sum_{j = 1}^M (u_{t,ij} - u_{t,ji}) \nonumber\\
&+ x_{t_0 t,i} + \sum_{t' = t_0}^{t} \sum_{j = 1}^M \mathbf{1}\left(\left\lfloor \frac{\tau_{t',ji}}{\Delta t} \right\rfloor=t-t'\right) \nonumber\\
& \times \mathbb{P}_{t'}(i|j)\ \text{min}(\bar{w}_{t'+1,j}, x_{t',j})\label{eq:dynamics_RHC}
\end{align}
\end{lemma}
Here the first term corresponds to ``leftover'' vehicles from time slot $t$, and the second term to the net number of idle vehicles dispatched to region $i$ at time $t$, i.e., right before the start of time slot $t + 1$.\footnote{For simplicity, we assume that dispatched vehicles are not assigned to any customers while traveling and that they always get to the destination regions in the next time slot, as we specify in (\ref{eq:opt_RHC}). Extending this definition still results in a linear optimization problem as in (\ref{eq:opt_RHC}).} The last two terms represent the vehicles that come into region $i$ at time $t$: the term $x_{t_0 t,i}$ corresponds to occupied vehicles at time $t_0$ that will drop off their passengers in time slot $t$. The final term corresponds to currently idle vehicles that will serve customers in the future and drop them off in the $i$-th region within time slot $t$. To derive this term, we sum over all regions $j$ and times $t'$ for which the expected travel time $\tau_{t',ij}$ to region $i$ places them in region $i$ at time $t$. The number of these trips given $j$ and $t'$ is then $\mathbb{P}_t(i|j)\text{min}(\bar{w}_{t'+1,j},x_{t',j})$, where $\mathbb{P}_t(i|j)$ is the fraction of trips that start at time $t$ in region $j$ and end in region $i$.

Assuming the $\bar{w}$ are known, we choose the dispatch actions $u_t$ so as to maximize the expected reward into the future:
\begin{proposition}\label{prop:RHC}
The optimal RHC policy $\left\{u_{t,ij}^\star | \forall t,i,j\right\}$ solves the linear optimization problem
\begin{equation}
\begin{aligned}\label{eq:opt_RHC}
& \underset{u_{t_0},...,u_{t_0+T}}{\text{maximize}}
& & \sum_{t=t_0}^{t_0+T} \gamma ^{t-t_0} r_t(u_t) \\
& \text{subject to}
& & \sum_{j = 1}^M u_{t,ij} \leq x_{t,i}; \ t = t_0, \ldots, t_0+T, \,  \forall i \\
& & & u_{t,ij} = 0, \qquad \{i,j,t \mid \tau_{t,ij} > \Delta t\}\\
\end{aligned}
\end{equation}
\end{proposition}
The first constraint in (\ref{eq:opt_RHC}) ensures that the total number of vehicles dispatched from the $i$th region does not exceed the number of idle vehicles in the $i$th region. The second constraint ensures that we do not dispatch vehicles to regions with travel times that exceed $\Delta t$, ensuring that all dispatch movement completes within a time interval; as noted above, this constraint may be relaxed without changing the linearity of the optimization problem. Using the definition of $r_t$ in (\ref{eq:reward_RHC}) and the vehicle dynamics (\ref{eq:dynamics_RHC}), we see by inspection that (\ref{eq:opt_RHC}) can be written as a linear optimization problem. For simplicity, we assume that the $u_{t,ij}$ are continuous variables, as there are generally a large number of taxis to be dispatched; we can then solve (\ref{eq:opt_RHC}) efficiently with known linear programming methods.\footnote{We show in our simulations that even with this approximation, the RHC policy yields significant performance improvement.} We retain $u_{t_0}^\star$ to execute now, updating the future dispatch actions $u_{t_0+1}^\star, \ldots, u_{t_0+T}^\star$ by re-solving (\ref{eq:opt_RHC}) in each future timeslot as new information arrives.

Algorithm~\ref{alg:RHC} presents the RHC dispatch algorithm using Proposition~\ref{prop:RHC}. In addition to solving (\ref{eq:opt_RHC}), the algorithm predicts the input trip times $\tau_{t,ij}$ and destination distributions $\mathbb{P}_t\left(d | o\right)$ from historical trip data (cf. Section~\ref{sec:experiments}). It then assigns specific idle vehicles to fine-grained dispatch locations within each region, given the number of vehicles to be dispatched to each region $(u_{t,ij}^\star)$. 
For computational efficiency, we specify vehicle locations in a greedy manner. We define $\mathcal{L}_i$ as a set of locations $l$ within the $i$-th region, which satisfies:
\begin{equation}
\label{eqn:agg}
x_{t,i} = \sum _{l \in \mathcal{L}_i} x_t(l),\;w_{t,i} = \sum _{l \in \mathcal{L}_i} w_t(l),
\end{equation}
where $x_t(l)$ and $w_t(l)$ represent the number of available vehicles and requests at location $l$ respectively. The demand supply distribution mismatch at location $l$ is then given by:

\begin{equation}
\eta_t{(l)} = \frac{x_t(l)}{\sum_l x_t(l)} - \frac{w_t(l)}{\sum_l w_t(l)}
\end{equation}
For each dispatch $u_{t,ij}^\star$, we send vehicles from locations with a greater mismatch, i.e., higher $\eta _t(l)$, to those with lower $\eta _t(l)$.

\begin{algorithm}[t]
\caption{Receding Horizon Control (RHC) dispatch policy at time slot $t$.}
\small
\label{alg:RHC}
\SetKwInOut{Input}{Input}
\SetKwInOut{Output}{Output}
\Input{$X_{t(t:t+T)}, W_{t+1:t+T}, \eta_t, \mathcal{I}_t$}
\Output{dispatch solution}
Update trip time estimation table $\tau_{t}$\;
Update destination distribution table $P_t(d|o)$\;
Solve LP problem and get $u_t^\star$\;
\For{i = 1:M}{
\For{j = 1:M}{
\For{k = 1:$u_{t, ij}^\star$}{
Select vehicle $f_t^{(n)} \in \mathcal{I}_t$ located at $\argmax _{l\in \mathcal{L}_i, x(l)>0} \eta _t(l)$\; 
Select dispatch location $d_t^{(n)} = \argmin _{l\in \mathcal{L}_j} \eta _t(l)$\;
Add ($n, d_{t}^{(n)}$) to the dispatch solution\;
}
}
}
\end{algorithm}

%% file: DQN.tex
Our DQN policy learns the optimal dispatch actions for individual vehicles. To do so, we suppose that all idle vehicles sequentially decide where to go within a time slot $t$. Each vehicle's decision accounts for the current locations of all other vehicles, but does not anticipate their future actions. Since drivers have an app that updates with other drivers' actions in real time, and it is unlikely that drivers would make decisions at the exact same times, they would have access to this knowledge. We can thus express the DQN reward function for each vehicle $n$:  
\begin{equation*}
r_t^{(n)} = r(s_t^{(n)}, a_t^{(n)}) = \sum_{t'=t-\delta}^{t} \lambda b_{t'}^{(n)} - c_{t'}^{(n)}, 
\end{equation*}
where $r_t^{(n)}$ is the weighted sum of the number of rides the $n$th vehicle picks up at time $t$, $b_t^{(n)}$, and the total dispatch time $c_t^{(n)}$, analogous to the RHC reward (\ref{eq:reward_RHC}). Here $\delta$ is the action update cycle, or minimum time between dispatches sent to a given vehicle. The action $a_t^{(n)}$ represents the region to which the $n$-th vehicle should head. We limit the action space in the range of the dispatch cycle similar to the RHC. Note that this reward is \emph{not} an explicitly specified function of $a_t$: DQN's model-free approach means that the exact relationship between $a_t$ and $r_t$ will the \emph{learned} by the DQN algorithm.

We define the optimal action-value function for vehicle $n$ as the maximum expected return achievable by any policy $\pi_t = \left\{a_{t'}^{(n)} \mid t' > t\right\}$:
\begin{equation}
Q^*(s,a) = \max_{\pi} \mathbb{E}\left[\sum_{t'=t}^{\infty} \gamma ^{t'-t} r_{t'}^{(n)}|s_t^{(n)}=s,a_t^{(n)}=a,\pi_t\right],
\end{equation}
which satisfies the Bellman equation:
\begin{equation}
Q^*(s,a) = \mathbb{E}_{s'}[r_t+\gamma \max_{a'}Q^*(s',a')|s_t^{(n)}=s,a_t^{(n)}=a],
\end{equation}
where $\mathbb{E}_{s'}$ denotes the expectation with respect to the environment after time $s'$. Instead of using the full representation of the state $s_t$, we approximate $Q$ with a neural network. We use $\theta$ to denote the weights of this Q-network, which can be trained by updating the $\theta_i$ at each iteration $i$ to minimize the following loss function:
\begin{equation}
L_i(\theta _i) = \mathbb{E}_{s,a,r,s'}[(r_t+\gamma \max_{a'}Q(s',a';\theta_i^-)
- Q(s,a;\theta_i))^2]
\end{equation}

This function represents the mean-squared error in the Bellman equation, where the optimal target values are substituted with approximate target values $r_t+\gamma \max_{a'}Q(s',a';\theta_i^-)$, using parameters $\theta_i^-$ from some previous iteration. 

The dispatch algorithm for the DQN policy is shown in Algorithm~\ref{alg:DQN}. The input and output of the DQN policy are the same as for the RHC policy (Algorithm~\ref{alg:RHC}). An action for each vehicle is selected by taking the argmax of the Q-network output. Whenever the algorithm adds a dispatch order to the solution, we update $X_{t(t:t+T)}$ according to the selected action. This update enables subsequent vehicles to take other vehicles' actions into account; note, however, that decisions are still made myopically with respect to possible future decisions taken by other vehicles, limiting coordination between vehicles. As in the RHC algorithm, after determining dispatched regions, the DQN policy finds specific dispatch locations in a greedy manner using the demand-supply mismatch $\eta_t$.

\begin{algorithm}[t]
\caption{Deep Q-Network (DQN) dispatch policy.}\label{alg:DQN}
\small
\SetKwInOut{Input}{Input}
\SetKwInOut{Output}{Output}
\Input{{$X_{t(t:t+T)}, W_{t:t+T}, \eta_t, \mathcal{I}_t$}}
\Output{dispatch solution}
\For{$f_t^{(n)} \in \mathcal{I}_t$}
{
Create a feature vector $\phi_t^{(n)} = \phi(f_t^{(n)}, X_{t(t:t+T)}, W_{t:t+T})$\;
Select $a_t^{(n)} = \argmax_{a} Q(\phi_t^{(n)}, a; \theta)$\;
Select a destination region $i$ from action $a_t^{(n)}$\;
Select $d_{t}^{(n)} = \argmax_{l \in \mathcal{L}_i} \eta_t(l) $\;
Add ($n, d_{t}^{(n)}$) to the dispatch solution\;
Update $X_{t(t:t+T)}$ based on $a_t^{(n)}$\;
}
\end{algorithm}

%% file: experiments.tex
To realistically evaluate our RHC and DQN policies, we design and implement MOVI as a taxi fleet simulator based on 15.6 million taxi trip records from New York City~\cite{data}. We used Python and tensorflow~\cite{tensor} for the implementation. 

We base MOVI's implementation on NYC Taxi and Limousine Commission trip records from May and June 2016, including the pickup and drop-off dates/times, locations, and travel distances for each trip~\cite{data}. While each pickup location in this dataset represents where a passenger hailed a taxi on the street, we assume the distribution of demand is similar when the passenger uses a mobile application. We use 12.8 million trip records from May 2016 to train the simulator and 2.8 million trip records from June 2016 for testing.

\begin{figure}
\centering
\vspace{-0.1in}
\includegraphics[width=0.2\textwidth]{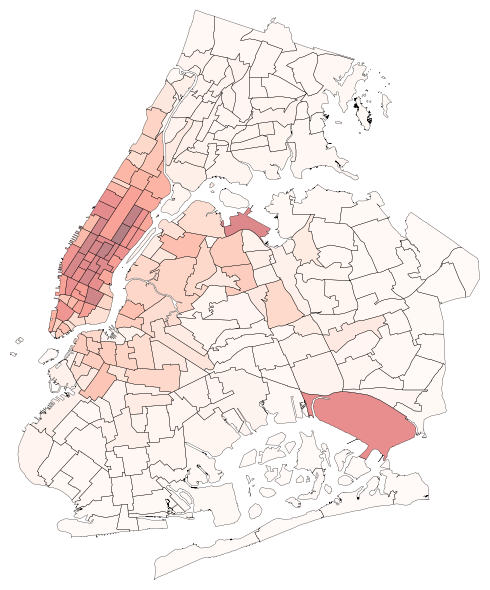}
\vspace{-0.1in}
\caption{Geographic demand distribution in NYC.}
\label{fig:demand}
\vspace{-0.05in}
\end{figure}

\begin{figure}
\centering
\includegraphics[width=0.48\textwidth]{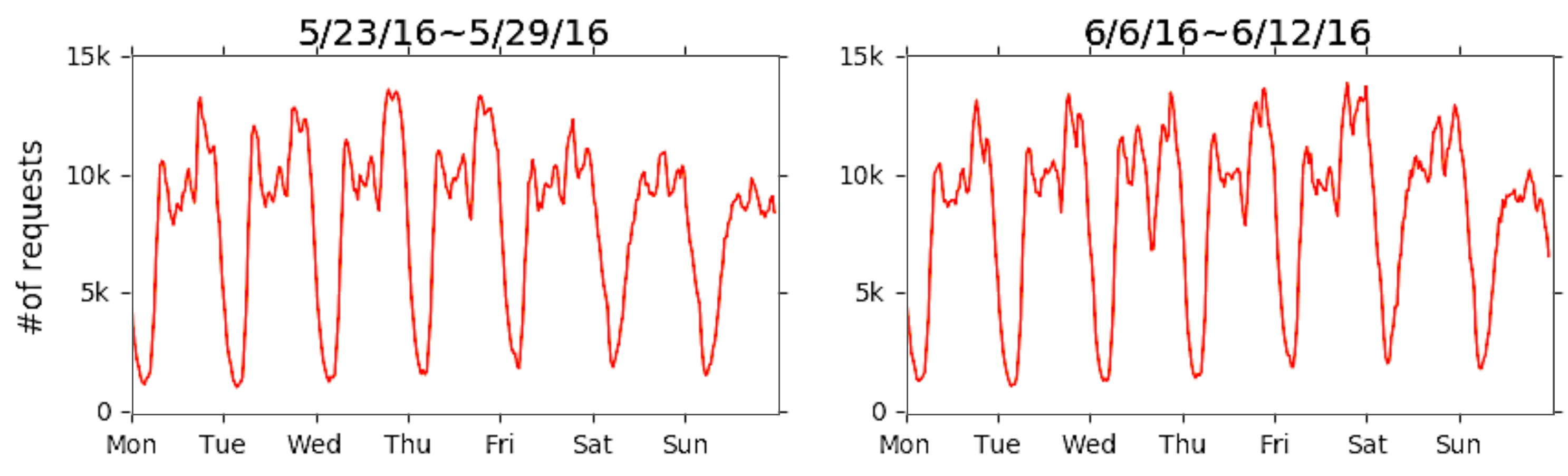}
\caption{Typical demand patterns in May and June 2016.}
\vspace{-0.05in}
\label{fig:demand2}
\end{figure}

We extract trip records in New York City within the area shown in Figure~\ref{fig:demand}'s heat map, which covers more than 95\% of trips in the dataset after removing records with outliers. The colored zones in the figure represent the number of total ride requests in the training data. The weekly numbers of ride requests for the training and test datasets are shown in Figure~\ref{fig:demand2}, indicating that the demand curves in both datasets exhibit the same daily periodicity, with a dip in demands over the weekend. In the discussion below, we outline the architecture of the simulator and then our RHC and DQN implementations. More details are given in~\cite{tr}.

\subsection{A Modular Architecture}
\begin{figure}[t]
\centering
\includegraphics[width=0.4\textwidth]{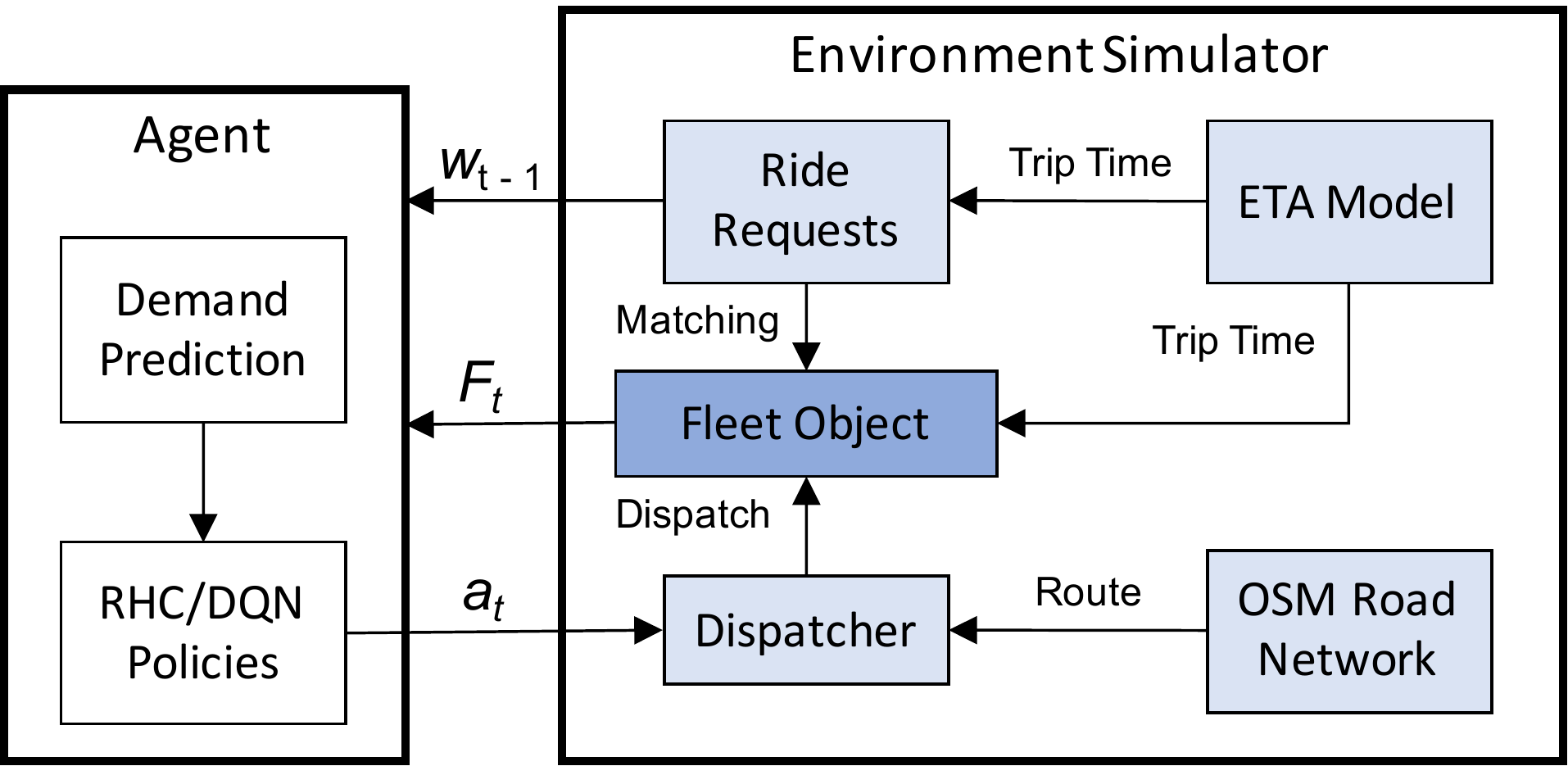}
\caption{MOVI's modular architecture. By separating the dispatch policy from the simulated environment, MOVI can compare the performance of our RHC and DQN policies.}
\label{fig:simulator}
\end{figure}

Figure~\ref{fig:simulator} presents MOVI's modular architecture design: to ensure a fair comparison between different dispatch policies, MOVI does not rely on the DQN policy. Instead, the dispatch policy is a separate module that does not affect the other simulator modules, which simulate policy responses in the surrounding environment. Thus, the simulated responses to dispatch decisions are policy-agnostic.

MOVI is based around the \textbf{fleet object}, which maintains the states $F_t$ of all vehicles at all times. In every time step, all vehicles update their states according to their matching and dispatch assignments. We discretize the city into $212 \times 219$ grid locations of size $150 \times 150$ m$^2$, which are later grouped into regions to compute the RHC and DQN dispatch policies. 
As detailed in Algorithm~\ref{alg:sim}, MOVI first initializes vehicles and generates ride requests based on the real trip records. The agent then computes the actions $a_t$, using either the RHC or DQN policy, and after the vehicles have gone to these locations, the dispatcher matches appropriate idle vehicles (those in the set $\mathcal{I}_t$) to requesting customers. When the agent sends a dispatch order to the vehicles, MOVI creates an estimated trajectory to the dispatched location based on the shortest path in the road network graph, and the vehicles move to dispatched locations within the trip time given by our ETA model. If there are no available resources in the customer's region, this ride request is rejected and disappears.

\begin{algorithm}[t]
\small
\caption{Fleet Simulator}\label{alg:sim}
Initialize the fleet state $F_0$\;
\For{$t=0:T_{max}$}
{
Load ride requests within time slot $t$\;
\For{each ride request}{
Select the closest vehicle $n$ to a pickup location\;
Compute dispatch trip time with the ETA model\;
Update $f_t^{(n)}$ according to the assigned ride request\;
}
Output $F_t, W_t$ to the agent\;
Get dispatch orders $a_t$ from the agent\;
\For{$n, d_t^{(n)} \in a_t$}{
Search the shortest path from the $n$-th vehicle location to dispatch location $d_t^{(n)}$\;
Compute dispatch trip time with the ETA model\;
Generate the future trajectory and assign it to the $n$-th vehicle\;
}
Update the fleet state $F_{t+1}$\;
}
\end{algorithm}


\textbf{Road Network Graph.}
We construct a directed graph to model the road network in the service area from Open Street Map data \cite{osm}. Whenever a vehicle is dispatched from an origin $o$ to destination $d$ for a vehicle, the simulator first finds the closest road edges to the $o$ and $d$ coordinates and then conducts an A* search for the shortest path between them.

\textbf{ETA (Estimated Time of Arrival) Model.}
To estimate the trip times $\tau^{(n)}_{l,m}$ for every dispatch $n$ at time $t$ from location $l$ to location $m$, we trained a multi layer perceptron. The input feature vector consists of the sine and cosine of the day of the week and hour of the day, pickup latitude and longitude, dropoff latitude and longitude, and trip distance. We use a random 70\% of the trip records in the training dataset to train the perceptron and 30\% for validation. With the trained model, the root-mean-square error (RMSE) for the training and validation datasets are 4.740 and 4.739 minutes respectively.

\textbf{Matching Algorithm.}
When a pickup request arrives, we assign it to the closest available vehicle. If there are no idle vehicles within five kilometers of the pickup location, the request is instead rejected. An assigned vehicle heads towards the pickup location with trip time $\tau$ predicted by the ETA model. After the pickup, the vehicle drives to the drop off location within the trip time on the actual trip record.

\subsection{Optimized Dispatch Policies}\label{sec:simulator_agent}

The agent in Figure~\ref{fig:simulator} runs either the RHC or DQN algorithm (Algorithms~\ref{alg:RHC} and~\ref{alg:DQN})). We detail our implementations of both in this section, after outlining the demand prediction that both algorithms use to represent the environment state.

\begin{figure}[t]
\centering
\vspace{-0.1in}
\includegraphics[width=0.35\textwidth]{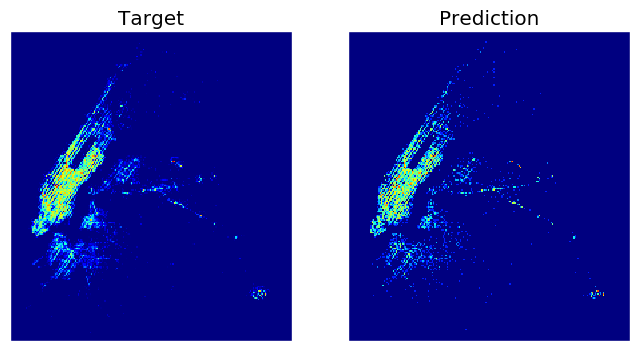}
\vspace{-0.05in}
\caption{Example of target and predicted demand heat maps.}
\label{fig:prediction}
\end{figure}

\textbf{Demand Prediction.}
To predict future demand, we build a small convolutional neural network. The output of the network is a $212 \times 219$ heat map image in which each pixel stands for the predicted number of ride requests in a given region in the next 30 minutes. The network inputs are the actual demand heat maps from the last two time steps; we capture daily periodicity by also including  the sine and cosine of the day of the week and hour of the day.
The network consists of two hidden layers; configuration details are in~\cite{tr}. We use thirty-minute timeslots, with the first 70\% of timeslots used for training and the last 30\% for validation. The RMSEs for the training and validation datasets are 1.047 and 0.980 respectively, i.e., our predicted demand is accurate to within a single request. Figure ~\ref{fig:prediction} shows an example of the target and predicted demand heat maps; they are visually identical.

\textbf{RHC Implementation.}
Since the RHC optimization involves $M^2 T$ variables, the number of regions significantly affects the computational time. Thus, we use the 226 taxi zones shown in Figure~\ref{fig:demand} as our dispatch regions. Predicted demand in each zone is calculated by aggregating outputs of the demand prediction model. We use a timeslot length of $\Delta t = 15$ minutes, reflecting the minute-scale runtime of the RHC algorithm, and $T=3$. 
The destination distribution given a trip's origin, $\mathbb{P}_t(d|o)$, was extracted from training data using a histogram count of the number of trips between regions for time $t$'s day of the week and hour of the day, thus taking into account cyclical demand patterns (cf. Figure~\ref{fig:demand2}).

\textbf{Streamlined DQN Training.}
For the DQN policy, we use smaller dispatch regions so as to utilize spatial convolution in training the $Q$ network. We divide the entire service area into a $43\times 44$ grid of regions, each of which is around $800 \times 800$ m$^2$. Each vehicle can move at most 7 regions horizontally or vertically from its current region, matching the constraint on vehicle travel times in the RHC optimization problem (\ref{eq:opt_RHC}) and resulting in a $15 \times 15$ map of possible destination regions. We select $\Delta t = 1$ minute as the length of each simulation step and a horizon of $T=30$, retraining the Q-network after each simulation step. Our technical report~\cite{tr} has more details on the Q-network input features and training.

\begin{figure}
\centering
\includegraphics[width=0.45\textwidth]{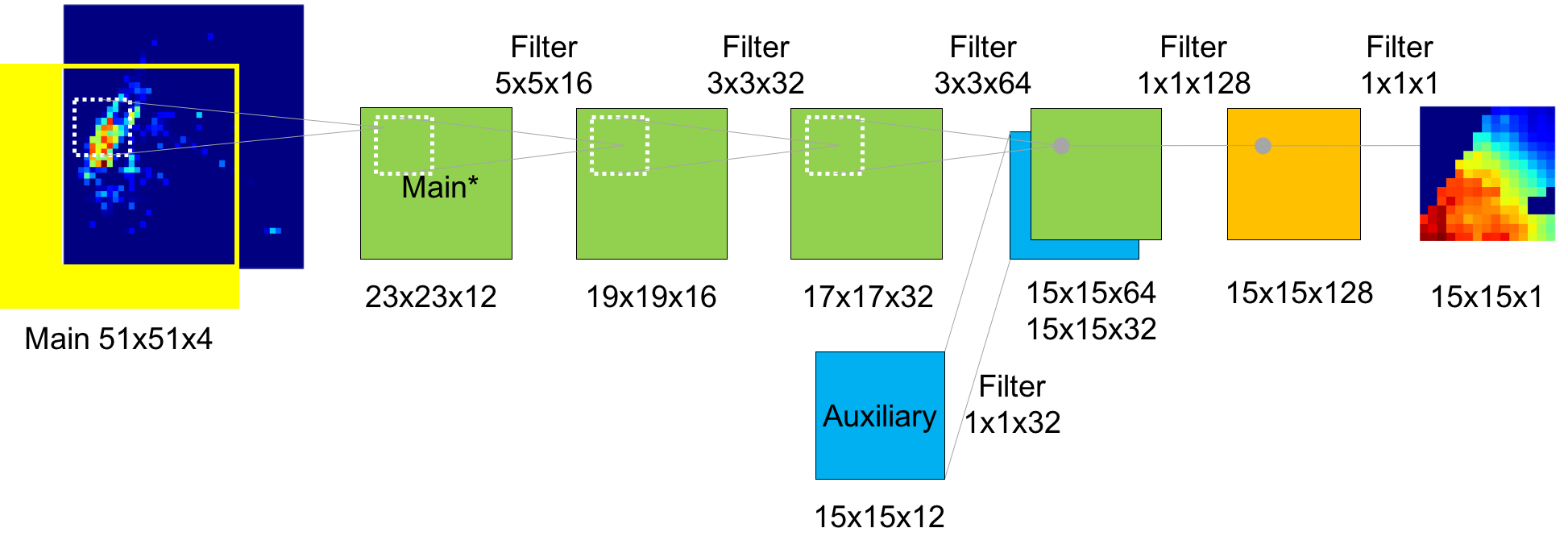}
\caption{Q-Network architecture.}
\label{fig:network}
\end{figure}

Figure~\ref{fig:network} presents our Q-network's architecture. We use a convolutional neural network with a $15 \times 15$ output map corresponding to the estimated Q-value for each possible action, given the input state. The input features are 
%
%
summarized in Table ~\ref{table:features}. In addition to the predicted ride requests $X_{t(t:t+T)}$ and future available vehicles $W_{t:t+T}$, we include environment features like the vehicle location, time of the day, and day of the week. 
We use three hidden layers and one output layer.

\begin{table*}[t]
\caption{Input features used for the Q-network. All are represented as planar images (cf. Figure~\ref{fig:network}.}
\vspace{-0.1in}
\label{table:features}
\centering
\begin{tabular}{|l|l|c|c|l|} \hline
Type & Feature & Plane size & \# of planes & Description \\ \hline
Main & Demand & $51\times51$ & 1 & Predicted number of ride requests next 30 minutes in each region \\
 & Supply & $51\times51$ & 3 & Expected number of available vehicles in each region in 0, 15 and 30 minutes \\
 & Idle & $51\times51$ & 1 & Number of vehicles in $\mathcal{I}_t$ in each region \\ \hline
Main* & Cropped & $23\times23$ & 5 & Main features applied (23, 23) cropping\\
 & Average & $23\times23$ & 5 & Main features applied (15, 15) average pooling with (1, 1) stride \\
 & Double Average & $23\times23$ & 5 & Main features applied (30, 30) average pooling with (1, 1) stride \\ \hline

Auxiliary & Day of week & $15\times15$ & 2 & Constant planes filled with sin and cos of the day of week \\
 & Hour of day & $15\times15$ & 2 & Constant planes filled with sin and cos of the hour of day \\
 & Position & $15\times15$ & 1 & A constant plane filled with 0 except current position of the vehicle with 1\\
 & Coordinate & $15\times15$ & 2 & Constant planes filled with current normalized coordinates of the vehicle \\
 & Move Coordinate & $15\times15$ & 2 & Normalized coordinate at this point \\
 & Distance & $15\times15$ & 1 & Normalized distance to this point from the center\\
 & Sensibleness & $15\times15$ & 1 & Whether a move at this point is legal\\ \hline

\end{tabular}
\vspace{-0.1in}
\end{table*}



Reinforcement learning is known to be unstable when a nonlinear approximator like a neural network is used to represent the $Q$ function. This instability is mainly due to correlations in the sequence of experiences and between the action-values $Q(s,a)$ and the target values $r+\gamma \max_{a'}Q(s',a')$. We use experience replay to remove these correlations and the Double DQN algorithm to prevent overestimation~\cite{schaul2015prioritized,DDQN}, using the RMSProp algorithm to train the Q-network. 

\begin{figure}
\centering
\includegraphics[width=0.5\textwidth]{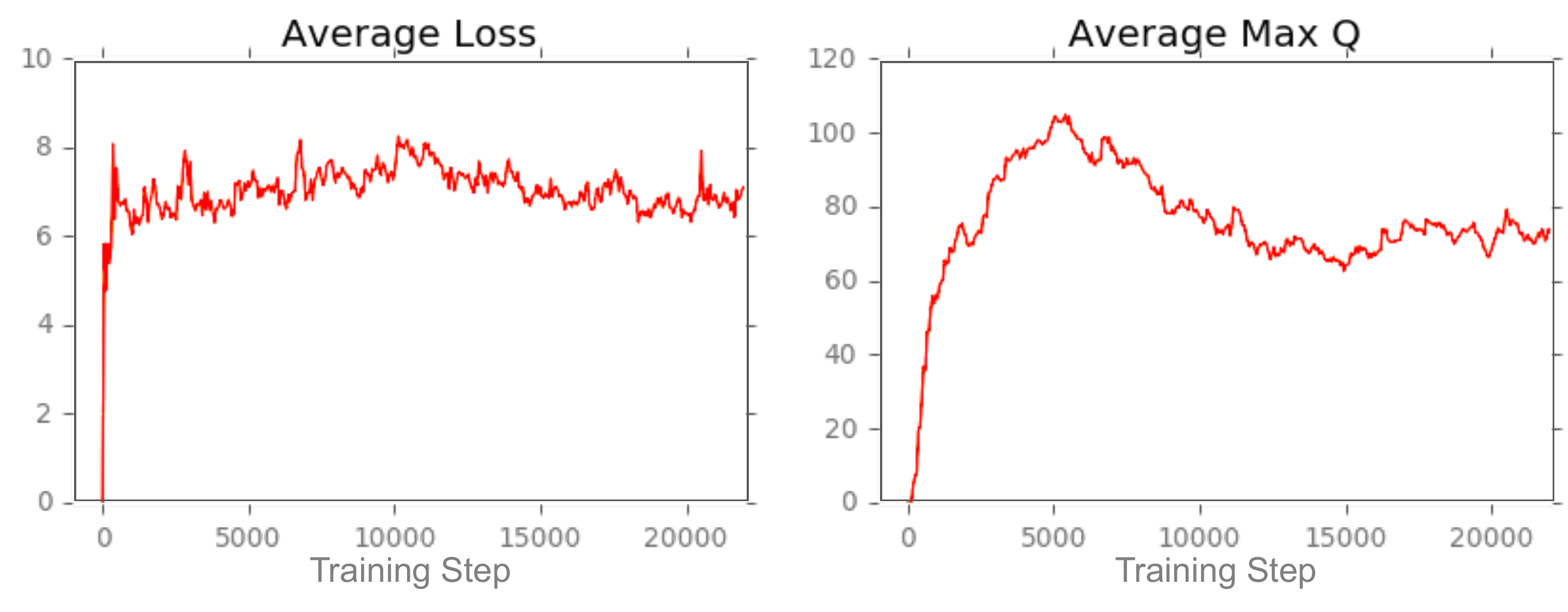}
\caption{Training curves tracking the agent's average loss and predicted action-value for the Q-network over the simulation.}
\label{fig:learning}
\end{figure}

We further streamline this training procedure to handle one of the biggest challenges in applying DQN to a fleet of vehicles: as vehicles execute policies, the state $s_t = \left(F_t, X_{t(t:t+T)}, W_{t(t:t+T)}\right)$ from the perspective of other vehicles changes, disrupting their Q-network training. Thus, we introduce a new parameter $\alpha$ as the probability that a vehicle moves in each simulation step, increasing $\alpha$ linearly from 0.3 to 1.0 over the first 5000 training steps. Thus, only 30\% of the vehicles move in the first step, which is roughly the percentage of vehicles taking actions in the optimal policy. We trained the Q-network for a total of 20,000 steps, corresponding to two weeks of data, and used a replay memory of the 10,000 most recent transitions. As illustrated in Figure ~\ref{fig:learning}, our method achieves stable loss and maximum Q-values over time. Once the average max-Q-value reaches 100, it starts decreasing: training in the previous time steps has improved taxis' Q-networks, allowing them to compete more for passengers and decreasing the average return an individual taxi can gain.

%% file: results.tex
For our evaluation, we use 2.8 million trip records from Monday, 6/6/2016 to Sunday, 6/12/2016. We assume that a day starts at 4 a.m. and ends at 4 a.m. in the following day, e.g., ``Monday'' is defined as 4 a.m. on Monday 6/6/2016 to 4 a.m. on Tuesday 6/7/2016. For each day, we conduct a dispatch simulation with 8000 taxi vehicles, whose initial locations are chosen from the pickup locations of the first 8000 ride requests in our data. We initialize the environment by first running the simulation for 30 minutes without dispatching.

For each day of the week, we compute three performance metrics: the average reject rate, wait time, and idle cruising time. The \emph{average reject rate} is defined as the number of rejected requests divided by the number of total requests in each day, and the \emph{wait time} is defined as the average time between a (un-rejected) pickup request originating to the time it is fulfilled. We define the \emph{idle cruising time} as the total driving time without passengers divided by the number of accepted requests. We also track the total trip time with passengers for each vehicle to compute the \emph{utilization rate}, or fraction of time for which a given vehicle is occupied.

\subsection{Performance Results}
\noindent We show the results of each policy before comparing them.

\begin{figure*}[t]
\centering
\begin{subfigure}{0.46\textwidth}
\includegraphics[width=0.96\textwidth]{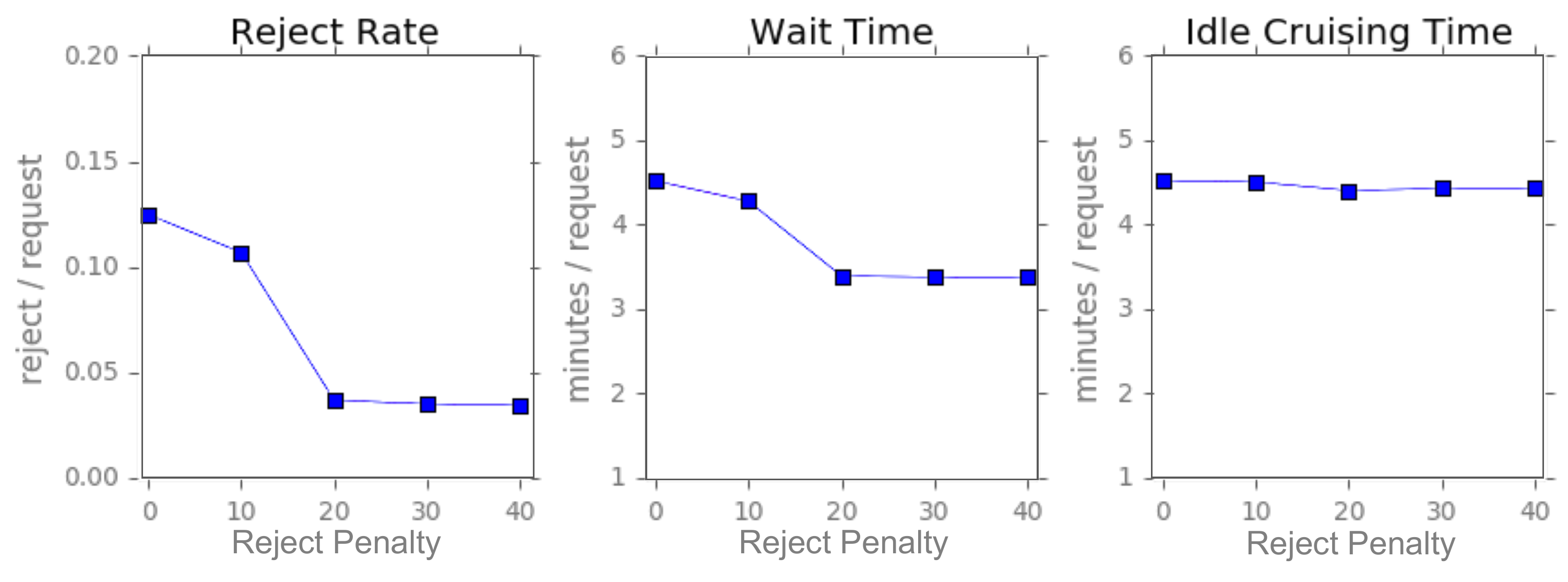}
\caption{Average performance vs. $\lambda$ for the RHC policy.}
\label{fig:rhc_lambda}
\end{subfigure}
%
\begin{subfigure}{0.46\textwidth}
\includegraphics[width=0.96\textwidth]{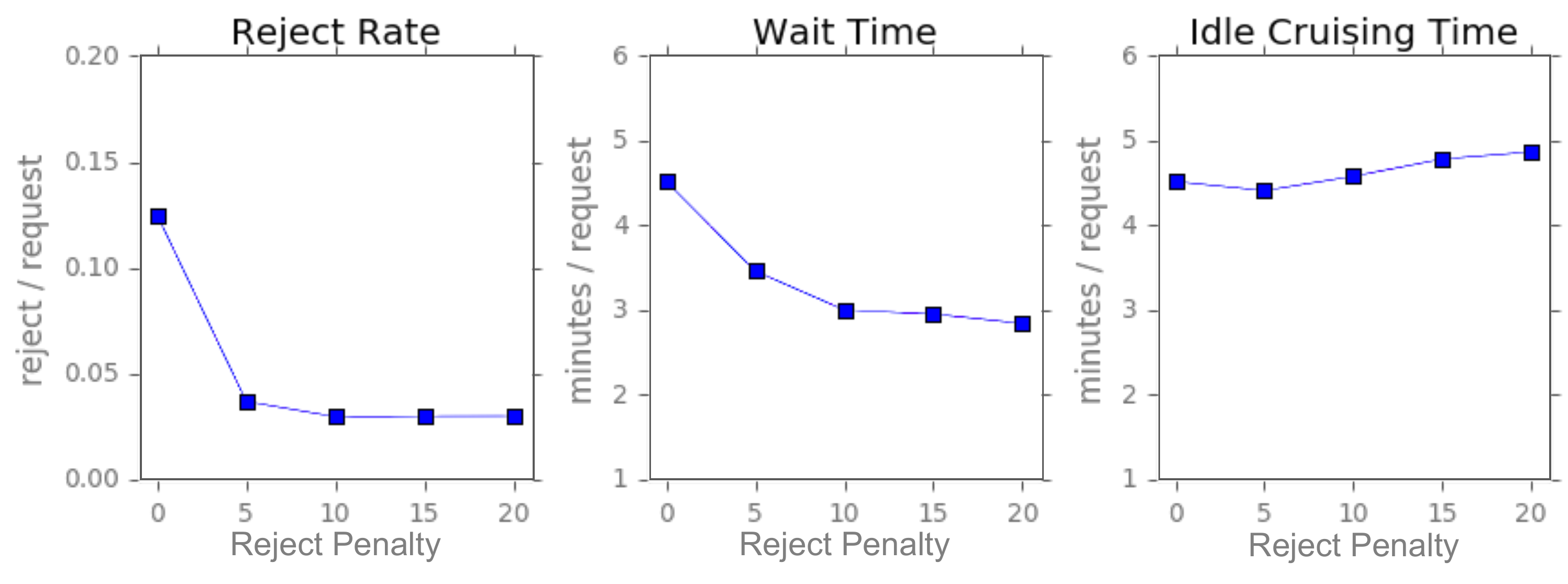}
\caption{Average performance vs. $\lambda$ for the DQN policy.}
\label{fig:dqn_lambda}
\end{subfigure}
\caption{The reject rate and passenger wait time improve as the reject penalty $\lambda$ increases, with little effect on the idle cruising time, in the (a) RHC and (b) DQN policies. The $x$-axis in all figures is the value of $\lambda$.}
\vspace{-0.1in}
\end{figure*}

\textbf{RHC Policy.} 
We conduct a simulation with the test dataset using the RHC policy and compute each metric's average value over a week. Figure~\ref{fig:rhc_lambda} shows the results with different reject penalties $\lambda$ from 0 to 40. While all three metrics improve as $\lambda$ increases from 10 to 20, they take nearly the same value when $\lambda \geq 20$. This result indicates that in practice, our three performance criteria do not conflict, which is surprising: we would expect the idle cruising time to increase as the reject rate decreases, due to vehicles traveling longer distances to pick up more passengers. The result suggests that most vehicles are close to passenger demand locations, yet many requests are not served quickly due to drivers not realizing this proximity. The floor on the reject rate as $\lambda$ increases, however, indicates that some requests are simply too far from any idle vehicles; our constraint on idle cruising time in (\ref{eq:opt_RHC}) then prevents any vehicles from traveling to their locations.

We also investigate the importance of maximum horizon $T$. In the technical report~\cite{tr}, we show that the performance does not change significantly with $T \geq 1$, indicating that there is limited value to coordinating vehicle locations too far into the future, perhaps due to limited ability to predict future demands.

\textbf{DQN Policy.} 
Similar to RHC, we evaluate our DQN policy by a simulation calculating each metric's average value over a week. Figure~\ref{fig:dqn_lambda} shows the results with different reject penalties $\lambda$ from 0 to 20. As seen in the figure, as $\lambda$ increases, the reject rate improves until $\lambda = 10$, while the idle cruising time increases modestly for $\lambda \geq 10$. As for the RHC policy, all metric values level off as $\lambda$ increases, indicating that there is a nonzero floor for the reject rate.

\textbf{Performance Comparison.}
\begin{figure}
\centering
\vspace{-0.1in}
\includegraphics[width=0.49\textwidth]{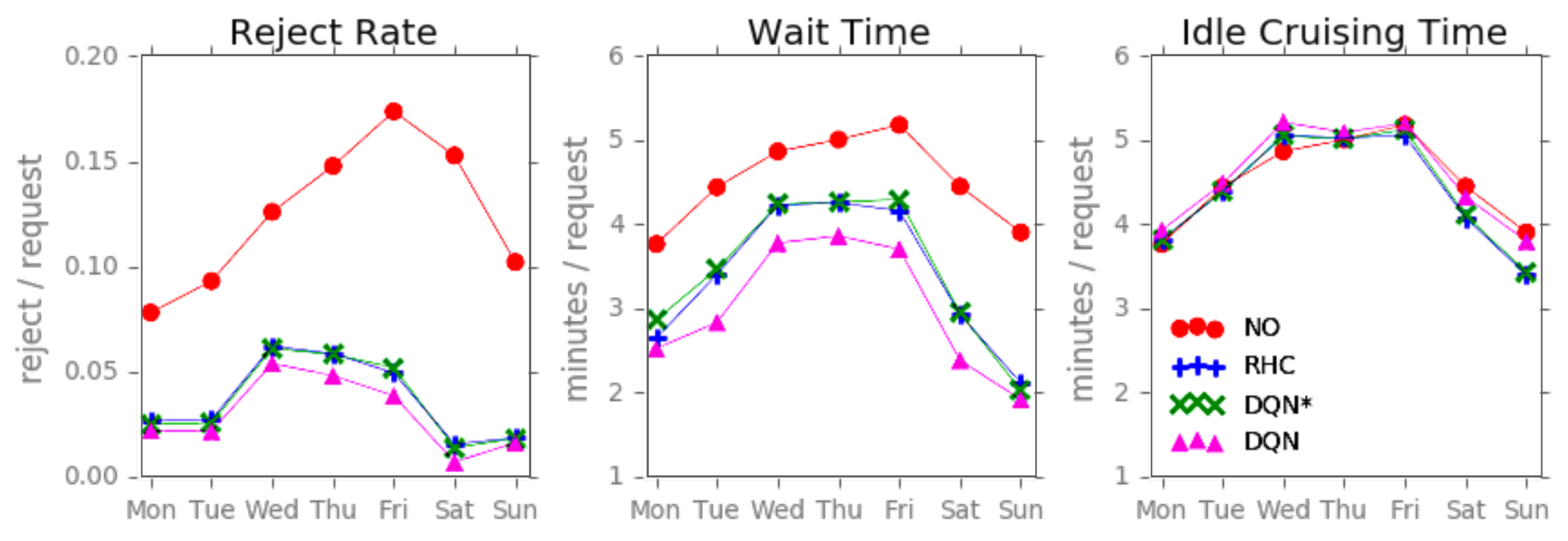}
\caption{DQN consistently outperforms RHC over one week.}
\vspace{-0.05in}
\label{fig:summary}
\end{figure}
We compare both dispatch policies to a simulation without dispatch. We summarize the results of no dispatch (NO), DQN with $\lambda = 10$ (DQN), and RHC with $\lambda = 20$ (RHC) in Figure~\ref{fig:summary}. Our results indicate that DQN outperforms RHC, but both significantly outperform no dispatching, indicating the value of optimized dispatch algorithms. They also suggest that DQN's better adaptability compensates for RHC's better coordination between vehicles.

In every day of the week, the RHC and DQN policies significantly reduce the reject rate and wait time compared to no dispatching, while the idle cruising time stays almost the same. The reject rate and average wait time of the DQN policy are reduced by 76\% and 34\% respectively compared with no dispatch, and by 20\% and 12\% compared with the results of the RHC policy. The idle cruising time of the DQN policy increases by 1.3\% compared with the time of no dispatch, and by 4.0\% compared with the time of RHC. Since DQN optimizes individual vehicle rewards, its policies may have individual drivers travel further to pickup requests, even though closer vehicles could also have served those requests.

\begin{figure}
\centering
\vspace{-0.1in}
\includegraphics[width=0.45\textwidth]{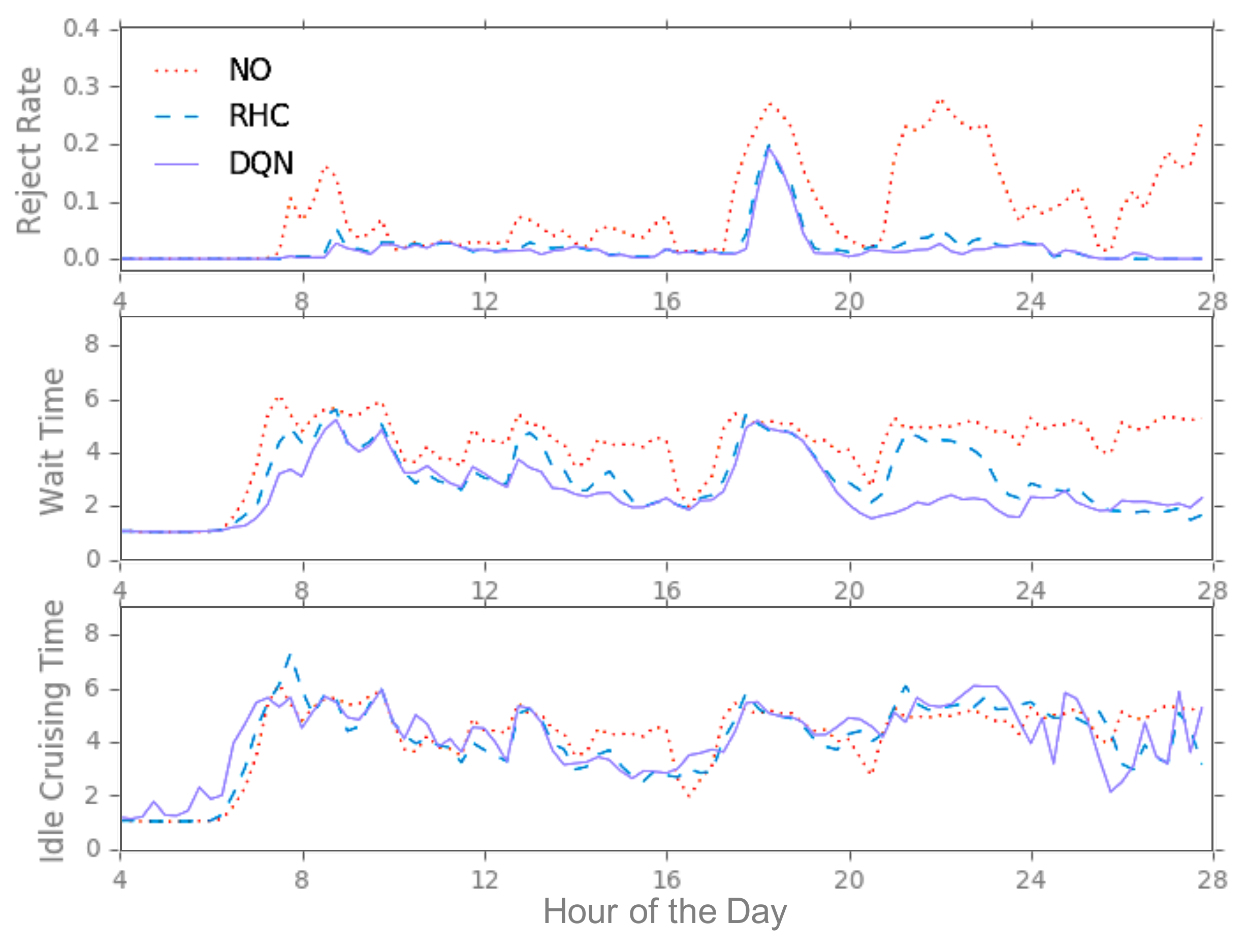}
\vspace{-0.1in}
\caption{DQN consistently outperforms RHC on Tuesday. The $x$-axis runs from 4 a.m. on Tuesday to 4 a.m. on Wednesday.}
\vspace{-0.1in}
\label{fig:tue}
\end{figure}

Figure~\ref{fig:tue} shows the reject rate, wait time, and idle cruising time with RHC and DQN dispatch and without dispatch on Tuesday. DQN dispatch consistently reduces the reject rate and wait time more than RHC; the technical report~\cite{tr} shows that this holds for Saturday as well. We note that the greatest reduction in the reject rate occurs at the time of highest demand, around 8pm to midnight (cf. Figure~\ref{fig:demand2}). Thus, \emph{optimized dispatch policies realize the most benefit at times of high demand}. At these times, without dispatching, drivers may not search for the locations of future ride requests, instead simply waiting for a request at their current locations. At these times DQN, but not RHC, drastically reduces passenger wait times, perhaps due to DQN having vehicles drive more to look for pickups. Indeed, the idle cruising times for the DQN policy are slightly higher than those for the RHC policy at these times, which is consistent with the overall results in~Figure~\ref{fig:summary}.

\begin{figure}
\centering
\includegraphics[width=0.48\textwidth]{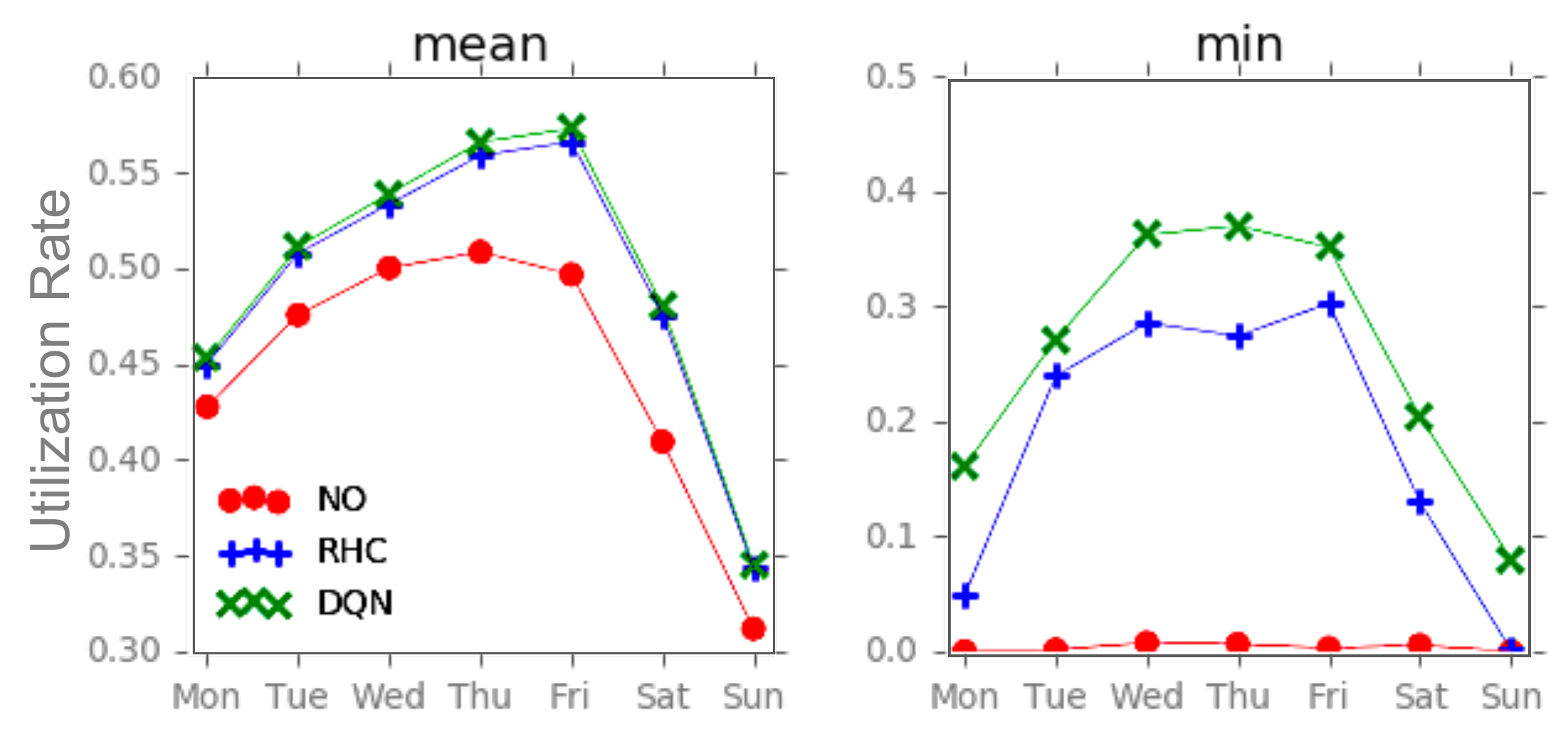}
\caption{Utilization rates vs. time. DQN has higher minimum but comparable mean utilization rates compared to RHC.}
\label{fig:utilization}
\end{figure}

We finally show that DQN more evenly distributes ride requests between vehicles by considering our 8000 vehicles' mean and minimum utilization rates in Figure~\ref{fig:utilization}. While the mean utilization rates for the two policies are almost the same, DQN's minimum utilization rate is much greater than RHC's. This smaller variance may be due to the fact that the DQN policy learns the optimal policy for an individual vehicle, meaning that every vehicle tries to take the best actions for itself. On the other hand, the RHC policy aims to maximize the total reward, forcing vehicles to take actions that may not benefit themselves, but do benefit the system as a whole. 

\subsection{DQN Advantages}
Despite the fact that the DQN policy does not make coordinated decisions for idle vehicles, our results show that DQN's reject rate is lower than RHC's on every day of week. We conjecture that this is due to DQN's much faster dispatch decisions, allowing the dispatch policies to rapidly adapt to the environment state. Compared to the fast computation of a neural network forward pass in each vehicle for DQN, which takes less than a hundred milliseconds, solving a linear program with tens of thousands of variables in the RHC policy is more expensive, taking from seconds to tens of seconds. 

In order to investigate the effect of the dispatch cycle, we simulate the DQN policy with the same dispatch cycle as the RHC policy. The results are plotted as DQN* in Figure~\ref{fig:summary}, showing that the reject rate is almost the same as RHC. DQN's faster on-demand dispatch thus helps the agent to adapt to disruptive environmental changes more quickly than RHC, at the expense of centralized cooperation. Even when DQN and RHC have the same dispatch cycles, DQN's lack of model constraints allows it to compensate for its lack of coordination and perform as well as RHC.

Obeying the DQN policy is generally more beneficial for drivers than the RHC policy, as DQN predicts the best action for each individual vehicle given its current state. Thus, the DQN policy may be more realistic to implement in real-world applications. Indeed, ride-sharing platforms like Uber allow drivers to choose where they go and which pickup requests to accept~\cite{uber-driver}, which may partially explain their success in improving passenger wait times compared to traditional taxi services.
Other potential advantages of a DQN approach include the fact that the same network architecture and input features can be used for different service areas; DQN's forward computation time is also independent of the number of dispatch regions, making it suitable for large service areas. In addition, other input features such as a vehicle's speed and capacity can easily be taken into account in the dispatch policies by simply adding them to the network input.

%% file: conclusions.tex
In this paper, we propose MOVI, a Deep Q-network (DQN) framework to dispatch taxis, which uses value-based function approximation with deep learning models and learns a optimal policy through directly interacting the environment. Dispatch simulation using taxi trip records in New York City shows that DQN policies lead to significantly fewer service rejects and wait times compared to no dispatching, outperforming the RHC policy with centralized coordination. In the future, it will be important to explore different network architectures and other input features such as the estimated time of each action to improve the DQN performance and computational efficiency, as well as establish a theoretical basis for DQN's superiority to RHC. Our work takes a first step in demonstrating the benefits of applying a model-free, practical  dispatch solution with state-of-the-art deep reinforcement learning techniques to large-scale taxi dispatch problems. 
